\documentclass[conference,a4paper]{IEEEtran}
\usepackage{spconf,amsmath,epsfig}

\usepackage{fancyhdr}

\def\L{{\cal L}}

\title{Detecting Cloud Presence in Satellite Images\\Using the RGB-based CLIP Vision-Language Model}

\name{Mikolaj Czerkawski, Robert Atkinson, Christos Tachtatzis\thanks{This work was funded by EPSRC under Grant EP/R513349/1.}}
\address{Department of Electronic and Electrical Engineering\\University of Strathclyde, Glasgow, UK}

\begin{document}

\fancyhf{}
\renewcommand{\headrulewidth}{0pt}
\fancyfoot[c]{}
\fancypagestyle{FirstPage}{
\lfoot{979-8-3503-2010-7/23/\$31.00 \copyright2023 IEEE}
\rfoot{IGARSS 2023}
}

\maketitle
\thispagestyle{FirstPage}

\begin{abstract}
    This work explores capabilities of the pre-trained CLIP vision-language model to identify satellite images affected by clouds. Several approaches to using the model to perform cloud presence detection are proposed and evaluated, including a purely zero-shot operation with text prompts and several fine-tuning approaches. Furthermore, the transferability of the methods across different datasets and sensor types (Sentinel-2 and Landsat-8) is tested. The results that CLIP can achieve non-trivial performance on the cloud presence detection task with apparent capability to generalise across sensing modalities and sensing bands. It is also found that a low-cost fine-tuning stage leads to a strong increase in true negative rate. The results demonstrate that the representations learned by the CLIP model can be useful for satellite image processing tasks involving clouds.
\end{abstract}

\section{Introduction}
\label{sec:intro}

    The text medium has long begun to play a prominent role in the processing of visual data over the last years, such as images~\cite{Radford2021}, or videos~\cite{Xu2021}. The use of language allows human users to easily adapt the computer vision models to their needs, which has prominently been used for purely creative purposes~\cite{Ramesh2021,Rombach_2022_CVPR,Saharia2022}, but also for zero-shot classification~\cite{Radford2021}. Vision-language foundations models could pave the way for many remote sensing applications that can be defined upon inference, without the need for extensive training or any training at all. This has been looked into in some works on visual question answering~\cite{Bazi2022,Lobry2020,Wei2023}, but the application to cloud presence detection remains unexplored.
    
    At the core of many text-based vision solutions stands CLIP, a vision-language model designed for measuring alignment between text and image inputs~\cite{Radford2021}. In this work, the capability of the CLIP model to recognize cloud-affected satellite images is investigated. The CLIP model operates on RGB images, while a typical solution to detect clouds in satellite imagery involves more than the RGB visible bands, such as infrared, and is often sensor-specific. Some approaches have explored the potential of an RGB-only cloud detection model~\cite{Ozkan2018}, but the task is considered significantly more challenging. Furthermore, the CLIP model has been trained on the general WebImageText dataset~\cite{Radford2021}, so it is not immediately clear how well it could perform with a task as specific as classification of cloud-affected satellite imagery.
    
    \begin{figure*}
        \centering
        \begin{tabular}{c c}
        \includegraphics[width=\columnwidth]{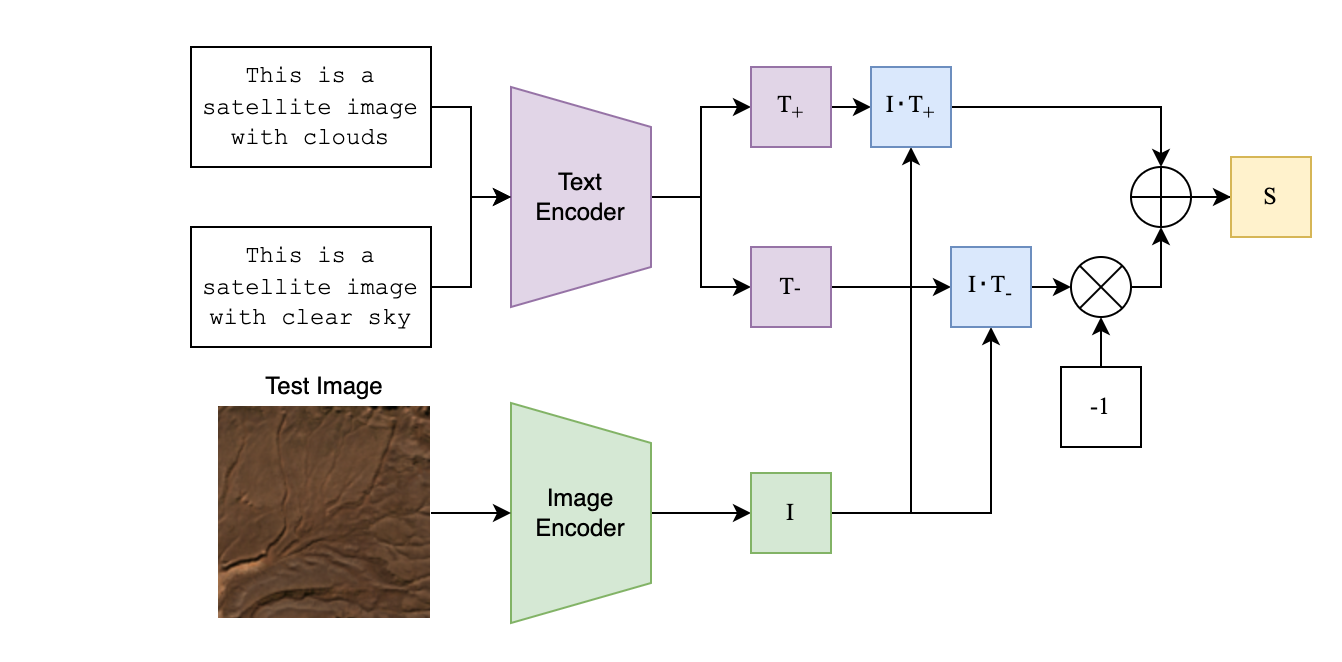} & \includegraphics[width=\columnwidth]{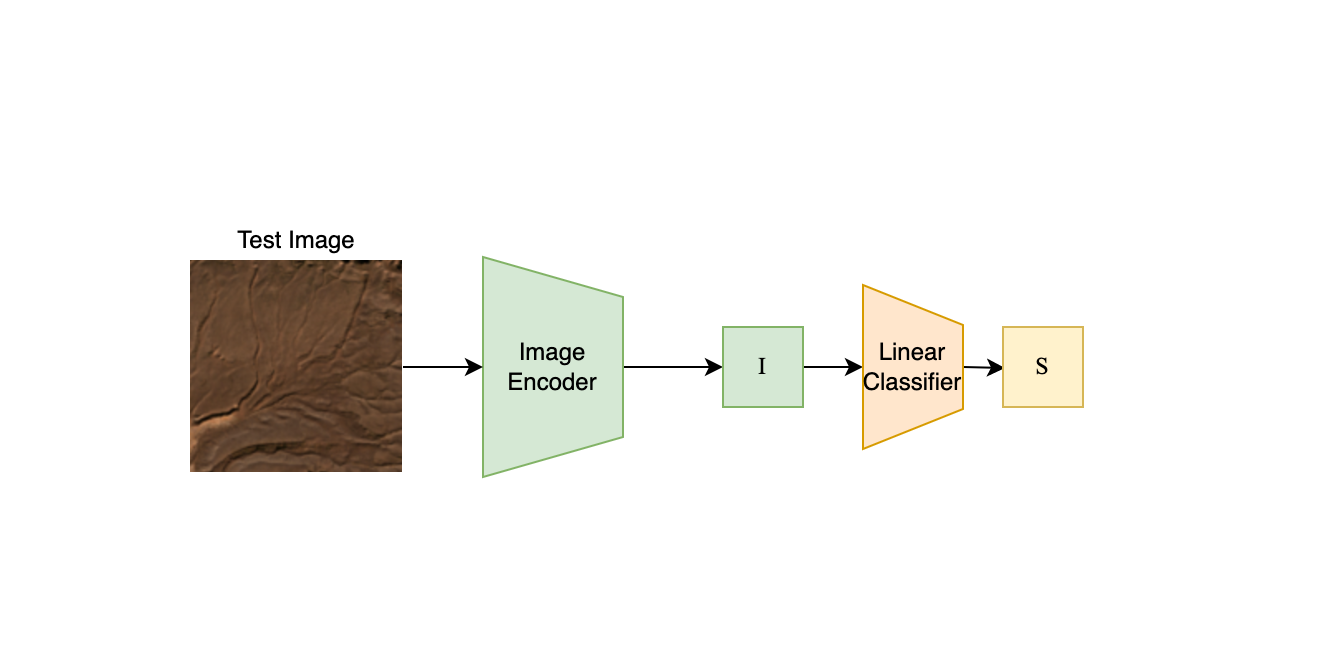} \\
        1. Text Prompts & 2. Linear Probe\\
        \includegraphics[width=\columnwidth]{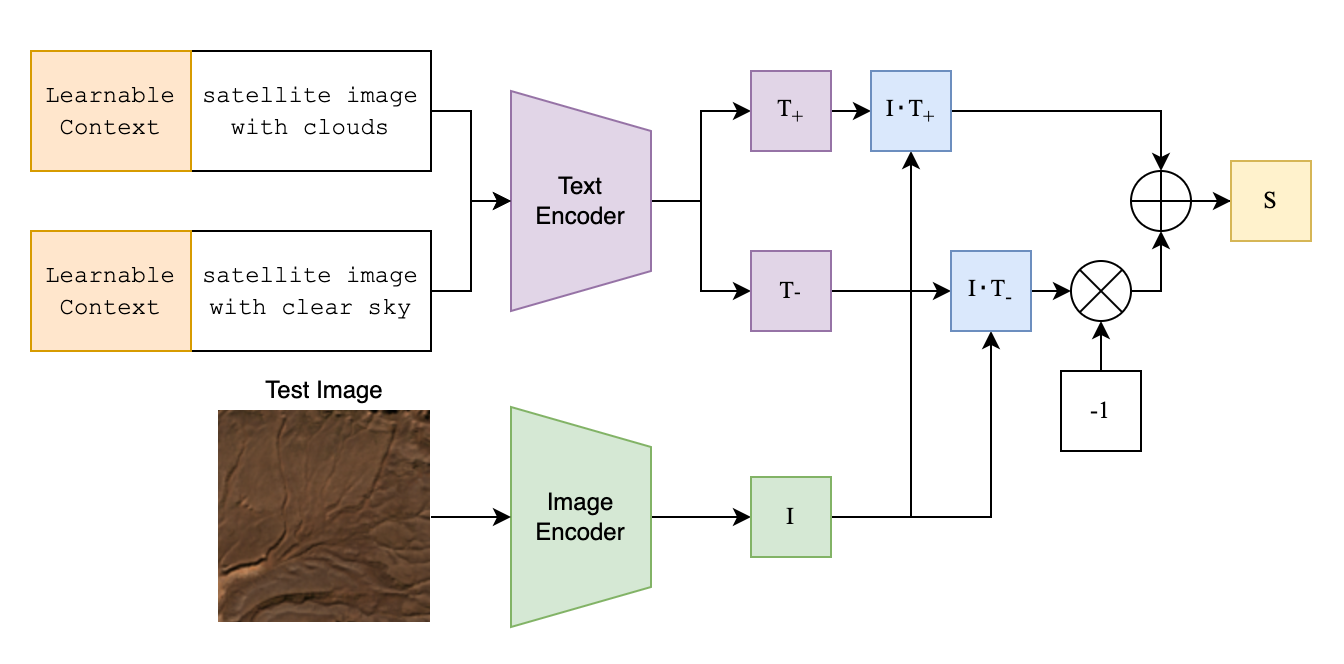} & \includegraphics[width=\columnwidth]{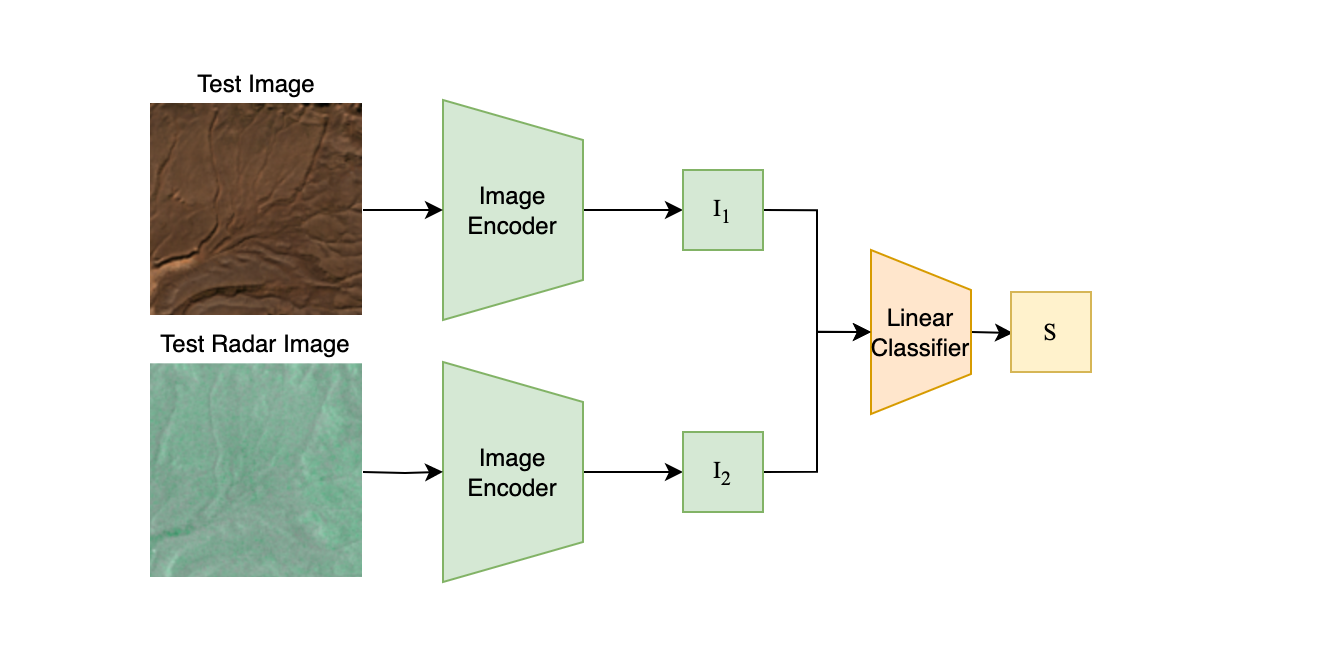}\\
        3. CoOp~\cite{Zhou2022} & 4. Radar
        \end{tabular}
        \caption{Diagrams of the proposed cloud presence detection methods based on the CLIP model}
        \label{fig:diagram}
    \end{figure*}
    
    In this work, the capability of the original pre-trained CLIP model (ViT-B/32 backbone) is tested. There are two important insights gained here: it allows to estimate the utility of representations learned by CLIP for cloud-oriented tasks (which can potentially lead to more complex uses such as segmentation or removal), and further, it can act as a tool for filtering datasets based on the presence of clouds.

\section{Method}
\label{sec:method}
    The CLIP model~\cite{Radford2021} has been designed for zero-shot classification of images where labels can be supplied (and hence, specified) as text upon inference. The CLIP model consists of separate encoders for text (transformer-based~\cite{Vaswani2017}) and images (either a Vision Transformer~\cite{dosovitskiy2021an} or a ResNet~\cite{He2016}) input, with jointly learned embedding space. A relative measure of alignment between a given text-image pair can be obtained by computing the cosine similarity between the encodings. This way, pair-wise similarity between any collection of images and text can be computed and compared, which enables use cases such as classification by extracting the label with the highest similarity to the input image.
    
    This work explores four variants of using CLIP for cloud presence detection, shown in Figure~\ref{fig:diagram}, one (fully zero-shot) based on text prompts (1), and (2)-(4) based on minor (1,000 gradient steps with batch size of 10, on only the training subset) fine-tuning of the high-level classifier module. The text prompts for method (1) were arbitrarily selected as \textit{``This is a satellite image with clouds''} and \textit{``This is a satellite image with clear sky''} with no attempt to improve them. In the case of (2), a linear layer is attached to the features encoded by the frozen image encoder. In the case of (3), a CoOp approach is employed, as described in~\cite{Zhou2022}. Finally, the Radar (4) approach applies a linear classifier to the image encodings of both RGB data and a false-color composite of the SAR Data (VV, VH, and mean of the two channels are encoded as 3 input channels).
    
\section{Evaluation}
\label{sec:evaluation}

    The approach is tested on two benchmark datasets: (i) CloudSEN12~\cite{Aybar2022}, containing Sentinel-2 and Sentinel-1 data and (ii) SPARCS~\cite{Hughes2019}, containing Landsat-8 imagery. Both datasets contain examples of cloudy images as well as images with no clouds present, representing the two classes in the cloud presence detection problem. This provides insight on the transferability of the tested methods. Since CLIP has never been trained to perform the task of cloud presence detection, it is not clear whether any CLIP-based method works well with any specific image modality. The ability of these techniques to transfer across sensing representations can indicate how powerful, or in other words, how generalisable a given solution is.
    
    The source sensor is not the only potential varying factor. Since the CLIP model is designed for RGB data, it accepts input with precisely 3 channels. Sentinel-2 and Landsat-8 contain multispectral data with more than 3 channels, including the channels approximating the RGB visible bands. Extracting the RGB channels would likely be the most straightforward approach of applying a pre-trained CLIP-based model to multi-spectral data. However, other bands are often useful too for the cloud presence detection task. For example, the annotators of the SPARCS dataset, while labeling the images, have been shown false-color images with bands B6 (SWIR), B5 (NIR), and B4 (Red) assigned to RGB channels, respectively~\cite{Hughes2019}. While these images do not correspond to the visible RGB data, they can still be interpreted by the CLIP model. To understand whether the same approach can be used on non-RGB images, two versions of the SPARCS dataset are tested here, one with the RGB bands and one with the false-color images observed by the annotators.
    
    The achieved performance is reported in Table~\ref{tab:detection_results} as true positive rate (TPR), the fraction of all cloudy images detected as cloudy; true negative rate (TNR), the fraction of all cloud-free images detected as cloud-free; and F1 score, a harmonic mean between the ratio of correct predictions among all cloudy samples and the ratio of correct predictions from all samples classified as cloudy. 
    
    \begin{table*}
        \centering
        \caption{Performance on cloud presence detection for the tested datasets and detection methods. The reported metrics include true positive rate (TPR), true negative rate (TNR) and F1 Score.}
        \vspace{5pt}
        \begin{tabular}{|l|ccc|ccc|ccc|}
            \hline
            Test Dataset & \multicolumn{3}{c|}{CloudSEN12} & \multicolumn{6}{c|}{SPARCS}\\
            Modality & \multicolumn{3}{c|}{S2/RGB}& \multicolumn{3}{c}{L8/RGB} & \multicolumn{3}{c|}{L8/B6-B4}\\
            \hline
            & TPR & TNR & F1 & TPR & TNR & F1 & TPR & TNR & F1 \\
            \hline
            1. Text Prompts & 0.929 & 0.638 & 0.919 & 0.922 & 0.737 & 0.907 & 0.900 & 0.737 & 0.895\\
            \hline
            \textit{Trained on:} & \multicolumn{9}{|c|}{S2/RGB}\\
            \hline
            2a. Linear Probe & 0.924 & 0.975 & 0.957 & 0.856 & 1.000 & 0.922 & 0.822 & 1.000 & 0.902\\
            3a. CoOp & 0.936 & 0.980 & 0.964 & 0.878 & 0.921 & 0.919 & 0.822 & 0.974 & 0.897\\
            4a. Radar & 0.930 & 0.960 & 0.959 & N/A & N/A & N/A & N/A & N/A & N/A\\
            \hline
            \textit{Trained on:} & \multicolumn{3}{c|}{L8/B6-B4} & \multicolumn{3}{c|}{L8/RGB} & \multicolumn{3}{c|}{L8/B6-B4}\\
            \hline
            2b. Linear Probe & 0.961 & 0.759 & 0.950 & 0.811 & 1.000 & 0.896 & 0.811 & 1.000 & 0.896\\
            3b. CoOp & 0.988 & 0.578 & 0.943 & 0.789 & 1.000 & 0.882 & 0.844 & 0.974 & 0.910\\
            \hline
        \end{tabular}
        \label{tab:detection_results}
    \end{table*}
    
    Three types of test data are used (corresponding to three sets of three columns in the table), starting with CloudSEN12 data with RGB Sentinel-2 input, and then Landsat-8 data from the SPARCS dataset with either RGB bands or B6-B4 false colour composite bands. The rows in the table correspond to different methods proposed in this work, including (1) zero-shot classification using text prompts, (2) linear probe fine-tuning, (3) CoOp~\cite{Zhou2022}, and (4) Radar-based input. Furthermore, the variants (2) and (3), fine-tuned on one type of 3-channel input can be tested on another source of 3-channel data from a different sensor. This is indicated by an additional letter, where (a) signifies models fine-tuned on Sentinel-2 RGB data, and (b) for models optimized on Landsat-8 data.
    
    The results in the first row with the zero-shot text prompt performance indicate that the CLIP-based model combined with the employed text prompts can achieve a high performance of at least 0.9 true positive rate, which means that the model can be quite reliable at picking up the cloudy samples. However, consistently across all three test datasets, the true negative rate is considerably lower, with values of 0.638 for Sentinel-2 data and 0.737 for Landsat-8 data (regardless of the representation), which means that more cloud-free images are classified as cloudy and that each technique exhibits a bias towards the positive label.
    
    The true negative rate is considerably improved by fine-tuning. For the CloudSEN12 dataset, the true negative rate increases to 0.975 for the linear probe approach (2a), 0.980 for the CoOp approach (3a) and 0.960 for the radar-based variant (4a). The true positive rate is consistently lower, meaning that some of the true positives are in result traded off for true negatives.
    
    For the models fine-tuned on the Landsat-8 data (2b. and 3b.), a similar effect is observed, with a very high true negative rate, and the true positive rate decreasing considerably from the level achieved in the fully zero-shot setting.
    
    The transferability is tested by applying the models from Sentinel-2 (2a-3a) to the SPARCS dataset and the models trained on Landsat-8 (2b-3b) to the Sentinel-2 images. In this case, the Sentinel-2 models appear to transfer better than the Landsat-8 models. However, a decrease in performance is observed upon transfer across modalities, especially in the case of transferring from Landsat-8 B6-B4 onto Sentinel-2 RGB data. This could mean that the discriminative relationships of the CLIP encodings for the false-colour data are quite different from the RGB data and do not transfer as well and this would need to be confirmed through further experimentation.

\section{Conclusion}
\label{sec:conclusion}
    The results presented herein demonstrate the potential of harnessing the general vision-language model of CLIP for processing clouds in satellite imagery with minimal training requirements.
    
    The CLIP model used in a zero-shot setting has been found to consistently struggle with detecting cloud-free images resulting in lower true negative rate than true positive rate. This weakness can be reduced by a low-cost training stage of only 1,000 optimization steps of a single linear layer trained on the CLIP image encodings. This approach leads to a high performance and observed transferability across tested sensor types and spectral bands.
    
    Lastly, an approach performing CLIP-based classification based on optical and radar data has been proposed and found to work at a comparable level to the optical-only approaches.
    
\bibliographystyle{IEEEbib}

\end{document}